\documentclass{article}

\PassOptionsToPackage{numbers, compress}{natbib}

\usepackage[preprint]{neurips_data_2024}

\usepackage[utf8]{inputenc} 
\usepackage[T1]{fontenc}    
\usepackage{url}            
\usepackage{booktabs}       
\usepackage{amsfonts}       
\usepackage{nicefrac}       
\usepackage{microtype}      
\usepackage{xcolor}         
\usepackage{multirow}
\usepackage{multicol}
\usepackage{graphicx}
\usepackage{hhline}
\usepackage{makecell}
\usepackage{subcaption}
\usepackage{amsmath}
\usepackage{mathtools}
\usepackage{ulem}
\usepackage{colortbl}
\usepackage{paralist}
\usepackage{bbding}
\usepackage{amssymb}
\usepackage{wrapfig}
\usepackage[linewidth=0.1pt]{mdframed}
\usepackage{multicol}
\usepackage{array}
\usepackage{geometry}
\usepackage{tablefootnote}
\usepackage{threeparttable}

\usepackage{subcaption, multirow, overpic, textpos}

\definecolor{green}{RGB}{0,150,10}
\definecolor{blue}{RGB}{0,148,181}
\definecolor{orange}{RGB}{194,153,107}
\usepackage[colorlinks,linkcolor=red,anchorcolor=green,citecolor=blue]{hyperref}

\makeatother
\newlength\savewidth
\newcommand{\tablestyle}[2]{\setlength{\tabcolsep}{#1}\renewcommand{\arraystretch}{#2}\centering\footnotesize}
\renewcommand{\paragraph}[1]{\vspace{1.25mm}\noindent\textbf{#1}}

\newcolumntype{x}[1]{>{\centering\arraybackslash}p{#1pt}}
\newcolumntype{y}[1]{>{\raggedright\arraybackslash}p{#1pt}}
\newcolumntype{z}[1]{>{\raggedleft\arraybackslash}p{#1pt}}

\usepackage[capitalize,noabbrev]{cleveref}

\title{DocGenome: An Open Large-scale Scientific Document Benchmark for Training and Testing Multi-modal Large Language Models}

\author{Renqiu Xia$^{1,2,*}$, Song Mao$^{1,*}$, Xiangchao Yan$^{1,*}$, Hongbin Zhou$^{1}$, \bf{Bo Zhang}$^{1,\ddagger}$ \\
\bf{Haoyang Peng$^1$}, \bf{Jiahao Pi$^1$}  \bf{Daocheng Fu}$^{1}$, \bf{Wenjie Wu}$^{1,2}$, \bf{Hancheng Ye}$^{1}$, \bf{Shiyang Feng}$^{4}$ \\ \bf{Bin Wang}$^{1}$, \bf{Chao Xu}$^{1}$, \bf{Conghui He}$^{1}$, \bf{Pinlong Cai}$^{1}$, \bf{Min Dou}$^1$, \bf{Botian Shi}$^{1,\ddagger}$ \\
\bf{Sheng Zhou}$^{3}$, \bf{Yongwei Wang}$^{3}$, \bf{Bin Wang}$^{4}$, \bf{Junchi Yan}$^{1,2}$, \bf{Fei Wu}$^{3}$, \bf{Yu Qiao}$^1$ \\[2.0mm]
$^1$ Shanghai Artificial Intelligence Laboratory, $^2$ Shanghai Jiao Tong University \\
$^3$ Zhejiang University, $^4$ Fudan University}

\begin{document}

\renewcommand{\thefootnote}{\fnsymbol{footnote}}
\footnotetext{*Equal contribution, 
$^\ddagger$Corresponding authors.}
\maketitle

\vspace{-0.15cm}
\begin{abstract}
\vspace{-0.30cm}

Scientific documents record research findings and valuable human knowledge, comprising a vast corpus of high-quality data. Leveraging multi-modality data extracted from these documents and assessing large models' abilities to handle scientific document-oriented tasks is therefore meaningful. Despite promising advancements, large models still perform poorly on multi-page scientific document extraction and understanding tasks, and their capacity to process within-document data formats such as charts and equations remains under-explored. To address these issues, we present DocGenome, a structured document benchmark constructed by annotating 500K scientific documents from 153 disciplines in the arXiv open-access community, using our custom auto-labeling pipeline. DocGenome features four key characteristics: \textit{1) Completeness}: It is the first dataset to structure data from all modalities including 13 layout attributes along with their \LaTeX\ source codes. \textit{2) Logicality}: It provides 6 logical relationships between different entities within each scientific document. \textit{3) Diversity}: It covers various document-oriented tasks, including document classification, visual grounding, document layout detection, document transformation, open-ended single-page QA and multi-page QA.  \textit{4) Correctness}: It undergoes rigorous quality control checks conducted by a specialized team. We conduct extensive experiments to demonstrate the advantages of DocGenome and objectively evaluate the performance of large models on our benchmark. DocGenome is available at \textcolor{teal}{\url{https://unimodal4reasoning.github.io/DocGenome_page}}

\end{abstract}
\vspace{-0.05cm}
\section{Introduction}
\vspace{-0.10cm}

Extracting data from scientific documents and developing large models to understand them is crucial for advancing AI-assisted scientific exploration and discovery~\citep{AlphaFold2021, AlphaFold-Multimer2021, baek2021accurate}. On one hand, scientific documents provide comprehensive, high-quality, logically rich corpora for training large models~\citep{lv2023kosmos, chen2023internvl, chen2024far, 2023gpt4V}. On the other hand, the ability of large models~\citep{lv2023kosmos, chen2023internvl, chen2024far, 2023gpt4V} to accurately understand scientific documents is considered as a crucial evaluation criterion.

However, we observed that current Multi-modal Large Language Models (MLLMs)~\citep{li2020docbank, zhong2019publaynet, pfitzmann2022doclaynet, da2023vision, wang2023cogvlm, chen2023internvl, chen2024far, bai2023qwen, alayrac2022flamingo, li2023monkey, tian2024mm, wang2024all, wang2024internvideo2, wu2023next, zhang2023gpt4roi, zhu2023minigpt} still struggle to understand the content of scientific documents as deeply as humans do. This challenge is primarily due to the inherently complicated multi-modal information present in scientific documents, such as multi-modal charts~\citep{xia2024chartx, wang2024charxiv}, intricate equations~\citep{wang2024unimernet, wang2024cdm}, and sophisticated logical relationships. Currently, MLLMs cannot effectively parse and comprehend such complicated modalities and logical relationships. To alleviate this challenge, we present DocGenome, an open large-scale scientific document benchmark constructed using the designed DocParser. 

\begin{figure*}[t]
\vspace{-0.65cm}
    \centering
    \resizebox{1\linewidth}{!}{\includegraphics{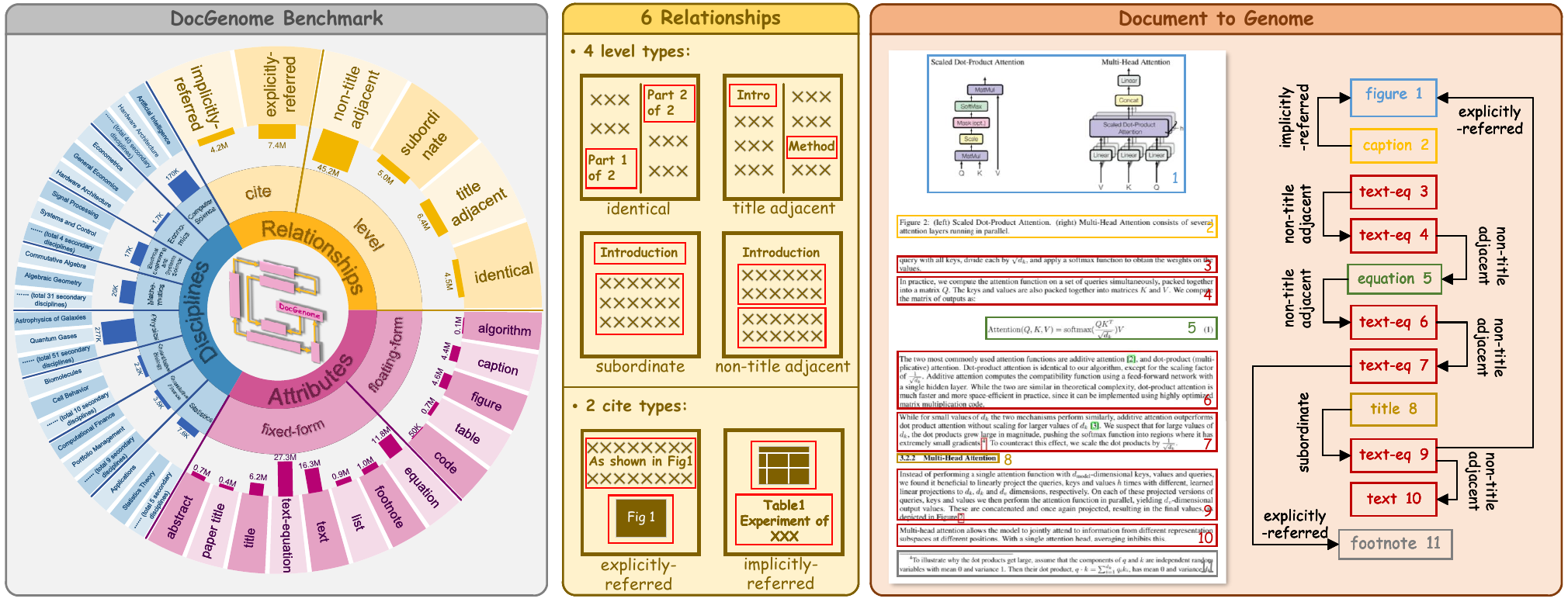}}
    \vspace{-0.35cm}
    \caption{\textbf{Overview of the DocGenome dataset.} Our work introduces DocGenome, a multi-modal dataset of academic documents encompassing 8 primary disciplines, 153 secondary disciplines, 13 categories of component units, and 6 types of entity relationships between units. We showcase an example of the paper~\citep{Vaswani2017AttentionIA} parsing into structured graph forms, termed as the document's genome, by leveraging the attributes and relationships of component units. } 
    \label{fig1:motivation}
\vspace{-0.15cm}
\end{figure*}

DocParser is a cutting-edge auto-labeling pipeline, which can generate both attribute information of component units and logical relationships between units by auto-annotating and structuring a large amount of unlabeled arXiv papers, with four stages: 1) data preprocessing, 2) unit segmentation, 3) attribute assignment and relation retrieval, and 4) color rendering as elaborated in Sec.~\ref{sec:auto_labeling_pipeline}. Furthermore, we utilize the proposed DocParser to label 500K scientific documents collected from the arXiv open-access community, and the resulting auto-annotated dataset is termed as DocGenome (illustrated in Fig.~\ref{fig1:motivation}), which contains 153 scientific disciplines and 7 document-oriented tasks including: document classification, visual grounding, open-ended single-page and multi-page QA tasks, document layout detection, Equation-to-\LaTeX\ transformation, Table-to-\LaTeX\ transformation, which is elaborated in Sec.~\ref{sec:eval_task}. Furthermore, we employ the quality grading and human validation methods to ensure the data quality as described in Sec.~\ref{sec:doc_analy} and Sec.~\ref{sec:qa_pair_quality_check}, respectively.

We conduct extensive experiments on the proposed DocGenome benchmark to objectively evaluate many mainstream MLLMs, including QWen-VL~\cite{bai2023qwen}, CogAgent~\cite{hong2023cogagent}, InternVL 1.5~\citep{chen2024far}, GPT-4V~\cite{2023gpt4V}, and \textit{etc}. The experiments on DocGenome also verify the effectiveness of the proposed dataset, demonstrating its ability to enhance the document understanding of the existing baseline models.

Our main contributions can be summarized as follows:
\begin{itemize}
    \item For the first time, we construct an open large-scale dataset that includes \textbf{500K} structured scientific documents with \textbf{13} categories of component units and \textbf{6} types of logical relationships between them. This dataset also encompasses various data types within scientific documents, such as Figure, Equation, Table, Algorithm, List, Code, Footnote, and \textit{etc}.
    \item To construct DocGenome, we design DocParser to automatically generate rich annotation information from the source code of a wealth of arXiv papers.
    \item DocGenome covers \textbf{7} document-oriented tasks, such as document layout detection, document transformation, multi-page QA, \textit{etc}. Besides, we conduct extensive verification and experiments based on these tasks to demonstrate that DocGenome can significantly enhance the document understanding capabilities of the existing baselines.
\end{itemize}
\section{Related Works}
\vspace{-0.10cm}

\begin{table*}[t]
\vspace{-0.20cm}
\centering
\small
\caption{Comparison with document-related benchmarks. `` - '' indicates that the corresponding part is not mentioned in the original paper. `` * '' means that each sample in their training set is cropped from the entire page, resulting in a total of 6.4M samples at the region level rather than the page level.}
\vspace{-0.15cm}
\resizebox{1.0\textwidth}{!}
{
\begin{tabular}{lcccccccc@{}}
\toprule
\multirow{2}{*}{Datasets}           & \multirow{2}{*}{\# Discipline}      & \# Category of   & \# Pages in & \# Pages in  & \# Task & \# Used Evaluation & Publication & With- \\ 
  &  & Component Units  & Train-set & Test-set  & Type  & Metric & Period & Entity Relation \\ 
\midrule
DocVQA~\citep{mathew2021docvqa}          & -                   &    N/A       &    11K      & 1K   &  1    &       2    &    1960-2000   & \XSolidBrush               \\
DocLayNet~\citep{pfitzmann2022doclaynet}          & -                   &    11       &    80K      & 8K   &  1    &       1    &    -   & \XSolidBrush               \\
DocBank~\citep{li2020docbank}          & -       &     13      &    0.45M      & \textbf{50K}   &       3   &  1       &    2014-2018   & \XSolidBrush               \\
PubLayNet~\citep{zhong2019publaynet} & -       &     5     &   0.34M     & 12K    &  1     &      1   &    -   & \XSolidBrush               \\
VRDU~\citep{wang2023vrdu} & -       &     10        &   7K      &   3K   &    3    &     1     &    -   & \XSolidBrush               \\
DUDE~\citep{van2023document} &    -    &  N/A          &    20K     &   6K   &    3     &    3     &   1860-2022    & \XSolidBrush               \\
$D^4LA$~\citep{da2023vision} & -       &  \textbf{27}          &    8K     &    2K   &    1    &    3      &    -   & \XSolidBrush               \\
Fox Benchmark~\citep{liu2024focus}          &  -                    &    5       &    N/A (No train-set)     & 0.2K  &     3      &       5      &  -   & \XSolidBrush               \\
ArXivCap~\citep{li2024multimodal}          &  32                    &   N/A       &    6.4M$^*$    & N/A    &      4    &       3      &  -   & \XSolidBrush               \\
\midrule
\textbf{DocGenome (ours)}     & \textbf{153}  & 13  & \textbf{6.8M} &  9K  & \textbf{7}  & \textbf{7} & 2007-2022 &\Checkmark  \\
\bottomrule
\end{tabular}}
\label{tab:related}
\vspace{-0.25cm}
\end{table*}

\noindent \textbf{Visual Document Datasets.} To comprehensively show the advantages of the proposed DocGenome dataset, we have reviewed visual document datasets and summarized them in Table~\ref{tab:related}. In earlier years, visual document datasets~\citep{li2020docbank, zhong2019publaynet, pfitzmann2022doclaynet, da2023vision} mainly aim to recognize the region categories of different regions from a given document, such as text region, table region, abstract region, and \textit{etc}. For example, DocBank~\citep{li2020docbank} constructs 500K high-quality document pages to enable the document layout model to utilize both textual and visual information. Recently, some research works~\citep{mathew2021docvqa, xia2023structchart, xia2024chartx, van2023document, li2024multimodal, liu2024focus} are proposed to build a document dataset with the enhanced diversity from multiple tasks, multiple modalities, and large-scale training data. By comparison, our DocGenome demonstrates more comprehensive features, including the number of disciplines and training samples covered, types of tasks, evaluation metrics, and entity relationships.

\noindent \textbf{Visual Document Understanding.} Research in the field of document Artificial Intelligence (AI) has made rapid progress, due to its successful applications in visual document layout analysis~\citep{wang2023docllm, van2023document, da2023vision, appalaraju2024docformerv2, luo2024layoutllm, huang2022layoutlmv3, he2023text2analysis} and image representation learning~\citep{zhou2024cross, he2022masked, Dosovitskiy2020AnII, bengio2013representation}. Inspired by Transformer~\citep{Vaswani2017AttentionIA}, LayoutLMv3~\citep{huang2022layoutlmv3} utilizes word-patch features to perform pre-training and designs a cross-modal alignment for document AI. UDIO~\citep{tang2023unifying} tries to unify multiple document-oriented vision tasks using task-specific prompting. Besides, Kosmos-2.5~\citep{lv2023kosmos} generates the text outputs by a shared decoder-only Transformer. mPLUG-DocOwl~\citep{ye2023mplug} boosts the OCR-free document understanding ability. Recently, ICL-D3IE~\citep{he2023icl} proposes an in-context-based learning framework to integrate LLM into document information extraction tasks and LayoutLLM~\citep{luo2024layoutllm} employs the layout instruction mechanism to improve the ability of document analysis.

\noindent \textbf{Multi-modal Large Language Models (MLLMs).} The development of MLLMs has profound impacts on the Artificial General Intelligence (AGI)  landscape. Recently, commercial MLLMs~\citep{2023gpt4V, team2023gemini, 2024claude, reid2024gemini} have experienced extremely rapid progress. GPT-4V~\citep{2023gpt4V} has significantly advanced the MLLMs. Google's Genimi series~\citep{team2023gemini, reid2024gemini} further enhance the ability of MLLMs to process text, images, and audio. Besides, open-source MLLMs~\citep{wang2023cogvlm, chen2023internvl, chen2024far, bai2023qwen, alayrac2022flamingo, lu2024not, li2023monkey, lin2024moe, liu2023controlllm, sun2023generative, tian2024mm, wang2024all, wang2024internvideo2, wu2023next, zhang2023gpt4roi, zhu2023minigpt} have also attracted great attention. Such MLLMs bring accessibility to the rapid development of AI, enabling widespread multi-modal applications and fostering innovation across industries.

\vspace{-0.10cm}
\section{Data Collection Methodology For DocGenome}

\subsection{Introduction of Auto-labeling Pipeline}
\label{sec:auto_labeling_pipeline}

\begin{figure*}[t]
\vspace{-0.55cm}
    \centering
    \resizebox{1\linewidth}{!}{\includegraphics{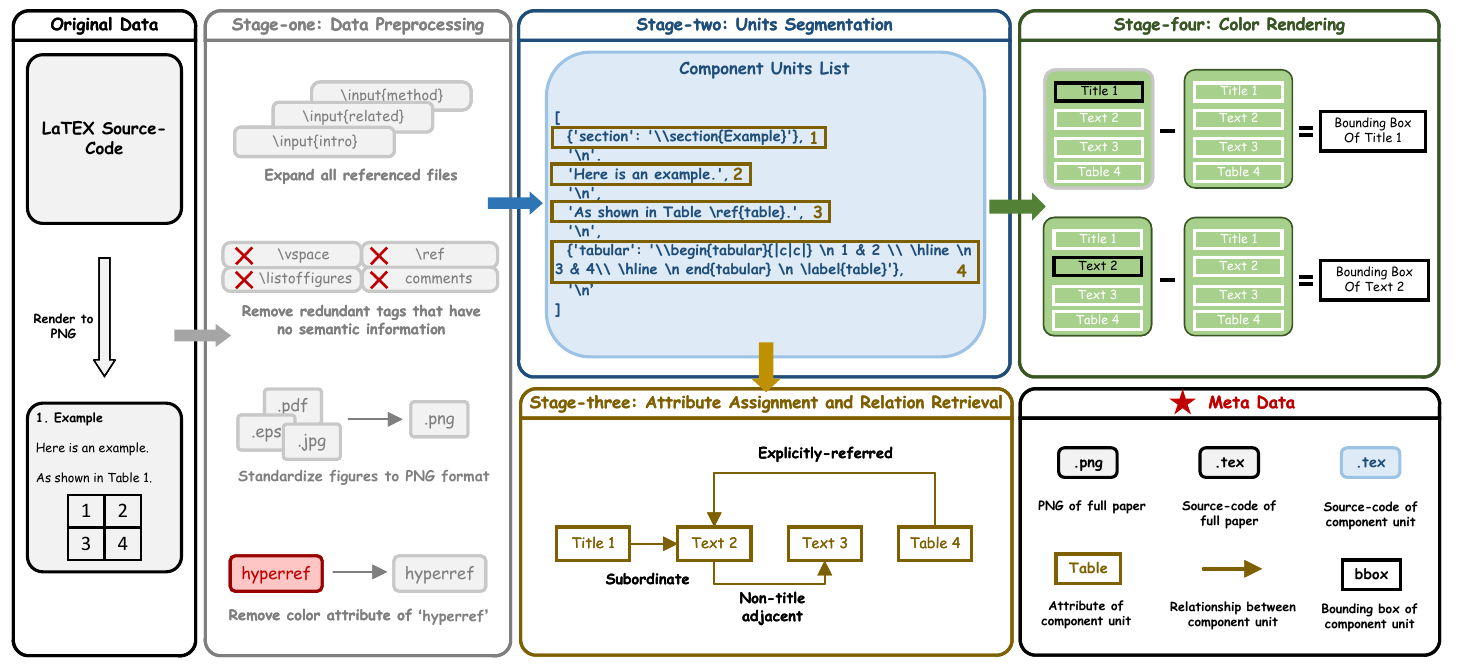}}
    \vspace{-0.40cm}
    \caption{\textbf{Schematic of the designed DocParser pipeline for automated document annotation.} The process is divided into four distinct stages: 1) Data Preprocessing, 2) Unit Segmentation, 3) Attribute Assignment and Relation Retrieval, and 4) Color Rendering. DocParser can convert \LaTeX\ source code of a complete document into annotations for component units with source code, attributes, relationships, and bounding box, as well as a rendered PNG of the entire document.} 
    \label{fig:pipeline}
\vspace{-0.25cm}
\end{figure*}

In this section, we present DocParser, a cutting-edge auto-labeling pipeline that streamlines the extraction of labeled source code from unlabeled arXiv data, serving as a key instrument for annotating the DocGenome dataset. As shown in Fig.~\ref{fig:pipeline}, the annotation process of DocParser is concisely divided into four stages, mitigating the issues of data scarcity and annotation expenses.

\noindent \textbf{Stage 1: Data Preprocessing.}
Our primary focus is to improve the data quality and enhance the compilation success rate of \LaTeX\ source code. 
Initially, we undertake an expansion of all files referenced by the \texttt{\textbackslash input} and \texttt{\textbackslash include} commands, followed by a series of crucial pre-processing steps. These steps encompass the integration of requisite environment packages, the exclusion of comment lines, and the removal of extraneous tokens such as \texttt{\textbackslash vspace}, \texttt{\textbackslash ref}, and other annotations that do not contribute to the semantic essence of the document.
Subsequently, we concentrate on standardizing the figure format within the \LaTeX\ source code, converting all graphical elements to the PNG format. 
Furthermore, we remove the color attribute from the ``hyperref'', ensuring that the \LaTeX\ source code is ready for targeted color rendering during annotation in stage 4.


\noindent \textbf{Stage 2: Units Segmentation.}
The objective of this phase is to automate the segmentation of content units, thereby streamlining the rendering process for distinct sections. We employ the \texttt{TexSoup}\footnote[5]{TextSoup package: \textcolor{teal}{\url{https://github.com/alvinwan/TexSoup}}.} library to decompose the \LaTeX\ source code into a structured list, delineating each individual component unit. This list is organized according to the reading order, ensuring a logical progression and facilitating the subsequent retrieval of relationships between the component units.


\noindent \textbf{Stage 3: Attribute Assignment and Relation Retrieval.} 
We have defined \textbf{13} fine-grained layout attributes (more details in Table~\ref{tab:layout_annotation} of Appendix~\ref{app:anno_explain}) for the component units decomposed in Stage 2, encompassing elements such as Algorithms, Captions, Equations, etc. For each unit, we match an appropriate attribute from the predefined set using keyword queries and regularization techniques to ensure a tailored and precise categorization. 
In the analysis of component unit relationships, units are categorized into two classes: \textbf{1) fixed-form units}, including Text, Title, Abstract, etc., which are characterized by sequential reading and hierarchical relationships readily discernible from the list obtained in Stage 2, and \textbf{2) floating-form units}, including Table, Figure, etc., which establish directional references to fixed-form units through commands like \texttt{\textbackslash ref} and \texttt{\textbackslash label}. The comprehensive set of \textbf{6} entity relationships is detailed in Table~\ref{tab:relationship_definitions}.
\begin{table}[t]
\small
\centering
\tablestyle{0.5pt}{1.1}
\setlength\tabcolsep{1pt}
\caption{The definition of logical relationships between component units.}
\begin{tabular}{y{70}y{170}y{160}}
\toprule
\textbf{Relation Name} & \textbf{Specific Description} & \textbf{Example} \\ \midrule
\textit{Identical} & Two units share the same source code. & Cross-column text; Cross-page text. \\
\textit{Title adjacent} & The two titles are adjacent. & (\textbackslash section\{introduction\}, \textbackslash section\{method\}) \\
\textit{Subordinate} & One unit is a subclass of another unit. & (\textbackslash section\{introduction\}, paragraph within Introduction) \\
\textit{Non-title adjacent} & The two text or equation units are adjacent.& (Paragraph 1, Paragraph 2) \\
\midrule
\textit{Explicitly-referred} & One unit refers to another unit via footnote, reference, etc. & (As shown in \textbackslash ref\{Fig: 5\} ..., Figure 5) \\
\textit{Implicitly-referred} & The caption unit refers to the corresponding float environment. & (Table Caption 1, Table 1) \\ \bottomrule
\end{tabular}
\label{tab:relationship_definitions}
\end{table}

\noindent \textbf{Stage 4: Color Rendering.}
The bounding box of a component unit is an additional label we aim to extract. After the segmentation phase in Stage 2, we render the target unit in black and all other units in white, to create two distinct PDFs. By performing a subtraction operation between these documents, we can obtain the detection box containing only the current unit, as illustrated in the top-right corner of Fig.~\ref{fig:pipeline}. For component units that traverse across hurdles or pages, we standardize the bounding box labels based on their unified source code information. This method effectively mitigates the issue where bounding boxes may be inadvertently divided, ensuring seamless and unified labeling for such units.

We automate the annotation process by sequentially applying DocParser's four stages and leveraging the complete \LaTeX\ source code. This yields not only the document's PDF but also the individual source code, bounding box, specific attributes for each component unit, and the relationships between units. Together, these elements constitute our DocGenome dataset.


\vspace{-0.10cm}
\subsection{DocGenome Benchmark Analyses}
\label{sec:doc_analy}
\vspace{-0.20cm}

Utilizing the DocParser automated annotation tool, we have annotated a corpus comprising 500K academic articles from the arXiv repository. 
Our analysis explores the diversity of the DocGenome benchmark, focusing on discipline distribution, content distribution, and quality grading.

\noindent \textbf{Discipline Distribution.}
The DocGenome consists of 8 primary disciplines, which collectively encompass 153 secondary disciplines\footnote[6]{According to the arXiv Category Taxonomy: \textcolor{teal}{\url{https://arxiv.org/category_taxonomy}}.}, reflecting a diverse and extensive coverage of academic research areas. The distribution across these disciplines is detailed in Fig. \ref{fig:secondary_distribution} of Appendix~\ref{app:distribution}.

\begin{figure}[t]
\vspace{-0.20cm}
\small
  \centering
  \resizebox{\linewidth}{!}{
  \begin{subfigure}[b]{0.41\textwidth}
    \includegraphics[width=\textwidth]{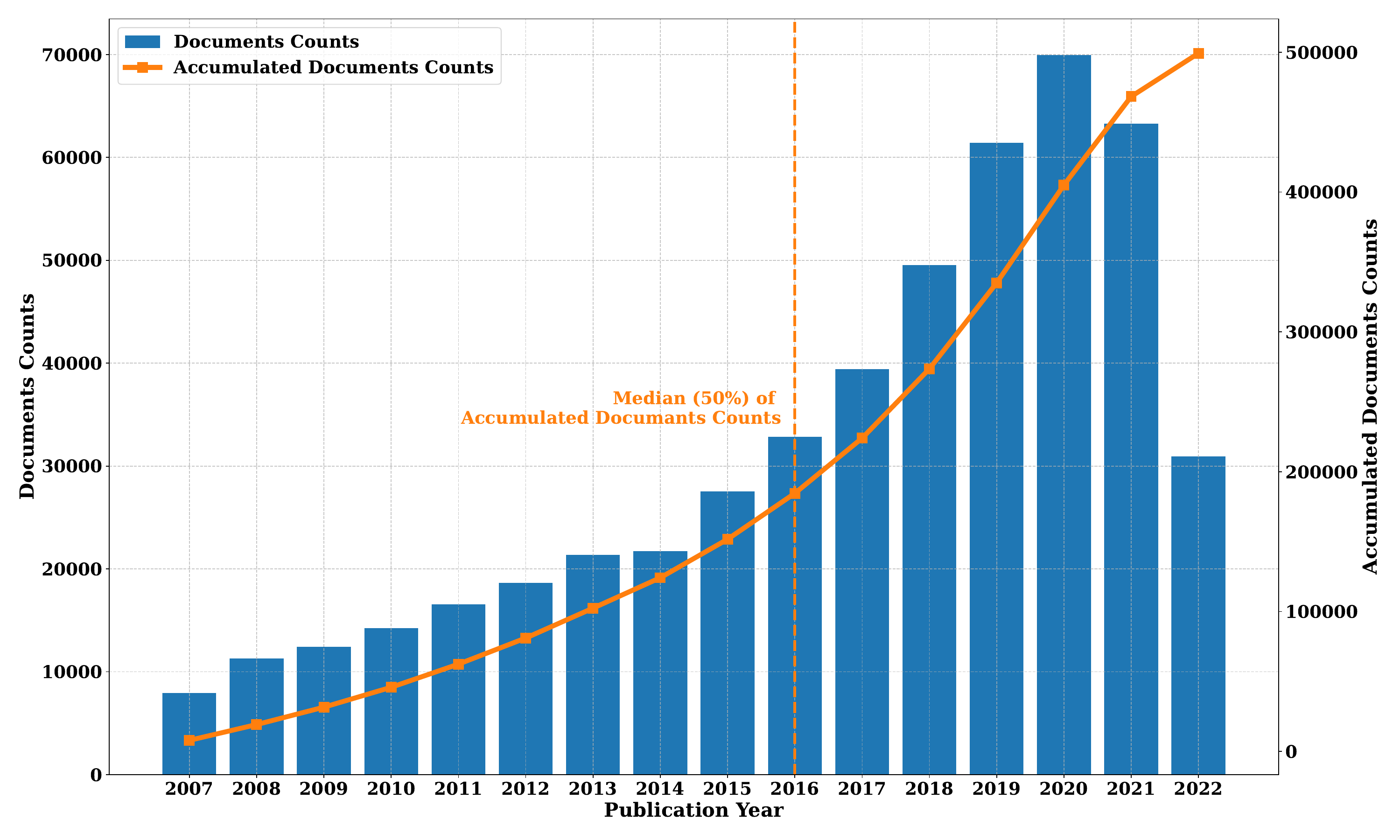}
    \vspace{-1.5em}
    \caption{}
    \label{fig:year}
  \end{subfigure}
  \begin{subfigure}[b]{0.41\textwidth}
    \includegraphics[width=\textwidth]{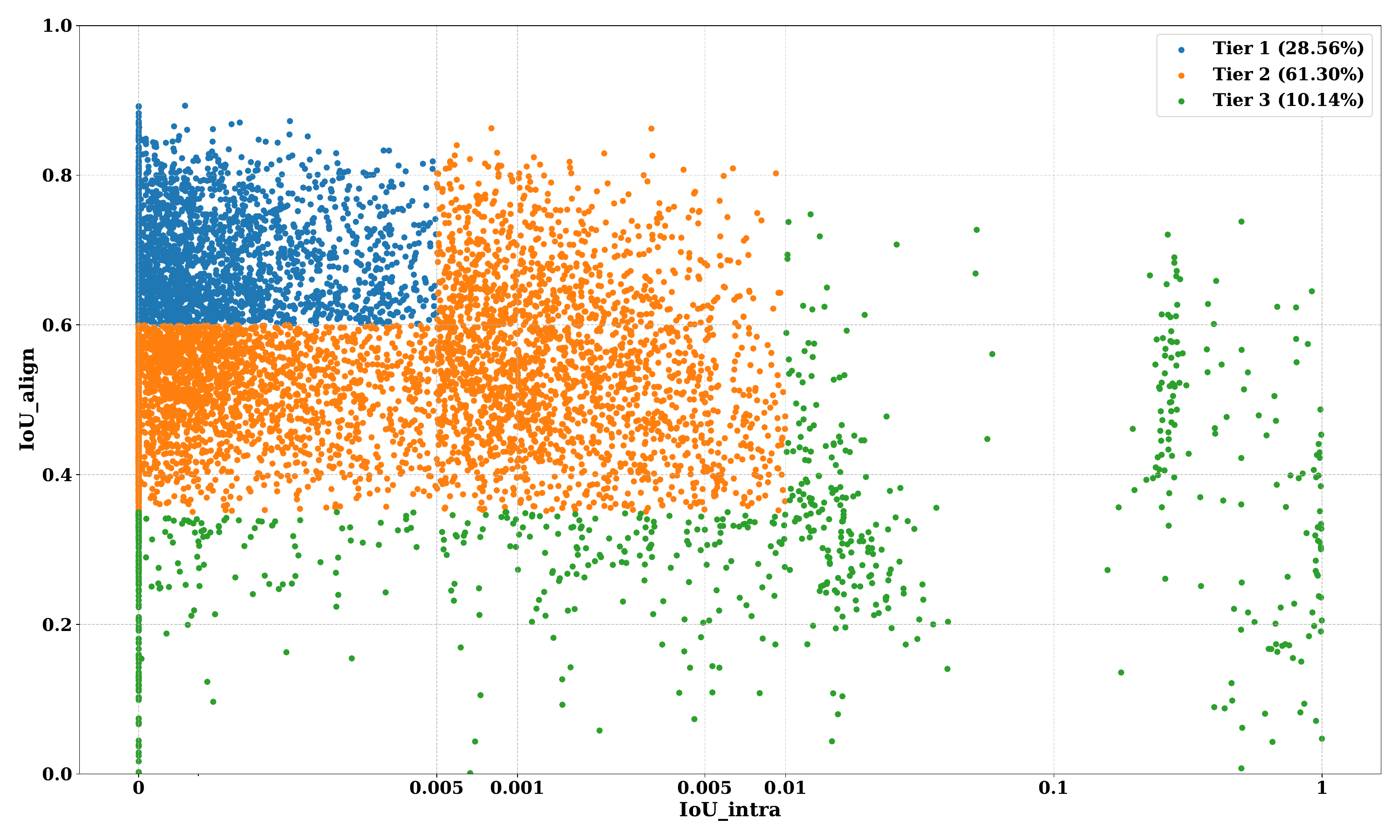}
    \vspace{-1.5em}
    \caption{}
    \label{fig:iou}
  \end{subfigure}}
  \vspace{-0.65cm}
  \caption{\textbf{Visualization of data distribution in DocGenome}. (a) Document publication counts over the years. (b) Distribution of three \textit{Tiers} determined by $IoU_{\text{intra}}$ and $IoU_{\text{align}}$.}
  \label{fig:docgenome_statistic}
\end{figure}

\noindent \textbf{Year Distribution.}
DocGenome archives articles from arXiv, ranging from 2007 to 2022, with a median publication year of 2016.  A significant portion, approximately 32.88\%, of these articles have been published since 2020. The distribution of these publications over time is depicted in Fig.~\ref{fig:year}.

\noindent \textbf{Content Distribution.}
We have examined two key aspects: the distribution of page counts and the labeling of component units. On the dimension of page counts, the dataset’s documents have an average page count of 13, with the longest document reaching 50 pages. The distribution of page counts is graphically represented in Fig. \ref{fig:page_distribution} of Appendix~\ref{app:anno_explain}.
Moving to the labeling perspective, we have annotated a substantial collection of 500K documents, totaling 74.5M component units and 68.5M relationship labels. In Fig.~\ref{fig1:motivation}, we present a detailed visualization of the distribution of both the attribute tags of the component units and the relationship labels. 

\noindent \textbf{Quality Grading.}
We establish two metrics to grade the data quality of the auto-labeled data that are generated using our DocParser. The first metric, designated as Eq.~\ref{eq:overlap}, measures the overlap among auto-annotated bounding boxes within each paper, thereby evaluating the intra-consistency of annotations:
\begin{equation}
\small
    IoU_{\text{intra}} = \frac{1}{N(N-1)} \sum_{i=1}^{N} \sum_{j=1, j \neq i}^{N} J(B_i, B_j),
    \label{eq:overlap}
\end{equation} 
where $J(B_i, B_j) = \frac{O(B_i, B_j)}{A(B_i) + A(B_j) - O(B_i, B_j)}$ is the $IoU$ between bounding boxes $B_i$ and $B_j$. $N$ is the total number of annotated bounding boxes in each paper. $O(B_i, B_j)$ represents the overlap area between bounding boxes $B_i$ and $B_j$. $A(\cdot)$ refers to the area of the bounding box.

Eq.~\ref{eq:filling} shows the second metric that quantifies the overlap between these annotated bounding boxes and the reference bounding boxes (predicted by DocXChain~\citep{yao2023docxchain}), providing an assessment of the annotations' alignment with established benchmarks, as formulated in Eq.~\ref{eq:filling}:
\begin{equation}
\small
    IoU_{\text{align}} = \frac{1}{N} \sum_{i=1}^{N} J(B_i, G_i),
    \label{eq:filling}
\end{equation} 
where $G_i$ is the $i$-th reference bounding box generated by DocXChain~\citep{yao2023docxchain}, $B_i$ refers to the bounding box that is closest to $G_i$ within our annotated ones.


A lower $IoU_\mathrm{intra}$ with a higher $IoU_\mathrm{align}$ indicates a higher quality of auto-annotated bounding boxes. Specifically, we split the collected paper into three tiers based on the annotation results. For the \textit{Tier}-1 set, we select the papers with $IoU_\mathrm{intra} < 0.05 \%$ and $IoU_\mathrm{align} > 60\%$, while those with $0.05\% \leq IoU_\mathrm{intra} < 1 \%$ and $IoU_\mathrm{align} > 35\%$ are packed in the \textit{Tier}-2 set, and the remaining papers are categorized as the \textit{Tier}-3 set. The distribution of three-tier data sets is shown in Fig.~\ref{fig:iou}, indicating that 28.56\% of the data was allocated to \textit{Tier}-1, 61.30\% to \textit{Tier}-2, and the other 10.14\% to \textit{Tier}-3.

\section{DocGenome-test: A Multi-task, Multi-modal, Comprehensive Evaluation Set for Document Understanding}
\label{test_set}

\subsection{Principles of Constructing Evaluation Set}

We use two principles to split the auto-annotated data into a high-quality evaluation set (\textbf{termed as DocGenome-test}) with precise annotation and a large-scale multi-modal training set (\textbf{termed as DocGenome-train}). First, the evaluation set should share the same discipline distribution as the collected data. Hence, the test data are uniformly sampled across each discipline. Second, the annotation of test data should be as precise as possible. Therefore, the test data are only sampled from the \textit{Tier}-1 set. Based on these two principles, we finally sampled 1,004 papers (covering 9K pages) as the test set from the overall 500K auto-annotated papers (containing 6.8M pages). As a result, the DocGenome-test covers 1,004 scientific documents with 1K document classification examples, 2K visual grounding examples, 3K QA pairs, 110K layout bounding boxes, 3K Table-\LaTeX\ pairs, and 5K Equation-\LaTeX\ pairs.

\subsection{QA Pair Generation and Quality Assurance}
\label{sec:qa_pair_quality_check}

In the DocGenome-test, we further design multiple Question-Answering (QA) pairs for each paper to comprehensively evaluate the document understanding capabilities of different models. For each paper sampler, two single-page QA pairs and two multi-page QA pairs are generated using GPT-4V~\citep{2023gpt4V}. Specifically, we instruct GPT-4V to randomly select two representative pages, extract useful information from the two pages respectively, and then generate corresponding single-page QA pairs. Additionally, we utilize GPT-4V to search for content-related paragraphs from different pages to construct the cross-page QA pairs, testing the model's ability to understand and integrate information across multiple pages. The QA pairs involve various commonly raised questions whose answers can be precisely inferred from the given paper. 

After generating QA pairs for all paper samples in the DocGenome-test, we invited professional faculty members from various fields to conduct the quality assurance checks. Each QA pair is reviewed by three reviewers for cross-verification. The first step involves the initial review by Kimi\footnote[8]{Kimi online API: \textcolor{teal}{\url{https://kimi.moonshot.cn}}.}, a well-known paper understanding model, to assess the initial correctness and identify the target location of QA information on the assigned page. Next, based on the provided location of QA information, two professional faculty members are assigned to manually and independently check each QA pair for accuracy, relevance, and clarity. At this stage, the quality evaluation involves the correctness, relevance, and rationality of the designed questions and the accuracy of the provided answer. Finally, the two manually-evaluated results, along with the automatically-evaluated result are cross-verified with the original text to ensure accuracy and consistency. Please refer to Appendix \ref{app:check_details} for more details.

\vspace{-0.15cm}
\subsection{Evaluation Tasks}
\vspace{-0.10cm}
\label{sec:eval_task}

To comprehensively evaluate the models' understanding capability of scientific documents, we design \textbf{7} tasks \textit{w.r.t} each paper document for the DocGenome-test, including document classification, visual grounding, open-ended single-page, and multi-page QA tasks, document layout detection, Equation-to-\LaTeX\ transformation, and Table-to-\LaTeX\ transformation. 

Specifically, document classification involves recognizing the field to which a paper belongs. Visual grounding involves identifying the content according to the provided visual components and textual prompts. Document layout detection refers to the localization and recognition of each layout block in given papers. Document transformation encompasses two format conversions, \textit{i.e.}, Table-to-\LaTeX\ and Equation-to-\LaTeX\ transformation. All tasks take the paper images as visual input for inference. The visual examples for each task are illustrated in Fig.~\ref{fig:visual_task} in Appendix~\ref{app:visual_task}.

\vspace{-0.20cm}
\subsection{Evaluation Metrics}
\vspace{-0.10cm}
\label{sec:eval_metric}

\noindent\textbf{Document Classification:} Top-1 Accuracy (\%) is used as the metric for document classification tasks, where higher values indicate better performance. 

\vspace{-0.10cm}
\noindent\textbf{Visual Grounding:} Edit Distance is used to evaluate the accuracy of visual grounding, with lower values indicating better performance.

\vspace{-0.10cm}
\noindent\textbf{Document Layout Detection:} mAP@0.5:0.95 is evaluated as the metric for document layout detection, where higher values indicate better performance.

\vspace{-0.10cm}
\noindent\textbf{Document Transformation:} We utilize Edit Distance, Jaccard Similarity, Cosine Similarity, and BLEU as metrics to comprehensively evaluate the document transformation task.


\vspace{-0.10cm}
\noindent\textbf{Open-ended QA:} GPT-acc (\%) is designed for tasks with open-ended answers, where outputs are evaluated against the ground truth using GPT-4. Please refer to Appendix~\ref{app:gpt_eval} for more details.

\vspace{-0.05cm}
\section{Experiments}
\vspace{-0.10cm}
\subsection{Compared Baselines and Implementation}
\vspace{-0.10cm}


\begin{table*}[!t]
\vspace{-0.25cm}
    \small
	\centering	
    \setlength\tabcolsep{4pt}
    \caption
	{
	\small	
        Comparison of state-of-the-art multi-modal large language models on the proposed DocGenome-test, including document classification, visual grounding, open-ended single-page, and multi-page QA tasks. Please refer to Sec.~\ref{sec:eval_metric} for the employed evaluation metrics.
	}
    \vspace{-0.15cm}
	\resizebox{1\textwidth}{!}{
	\begin{tabular}	{l l |c | c  c | c c}
	\toprule	 	
	 \multirow{3}{*}{Model} & \multirow{3}{*}{{\#Params}} & \multirow{2}{*}{Classification}  &   
        \multicolumn{2}{c} {Visual Grounding} & \multicolumn{2}{c} {Document QA} \\
        & &  & Title & Abstract & Single-Page & Multi-Page \\ 
	& & Acc$\uparrow$  &  Edit Distance$\downarrow$ & Edit Distance$\downarrow$ &  GPT-acc$\uparrow$ & GPT-acc$\uparrow$  \\
	\midrule  
    \multicolumn{2}{l}{\textit{~~\textbf{Multi-modal Large Language Models}}} \\
        QWen-VL~\cite{bai2023qwen} & 9.6B & 0.8237
	& 0.0775 & 0.8054 & 0.1156  & 0.0627 
	\\
        CogAgent~\cite{hong2023cogagent} & 17.3B & 0.5857
	& 0.0166  & 0.5306     & 0.1772  & - 
	\\    
        DocOwl-1.5~\cite{hu2024mplug} & 8.1B
	& 0.3307  & 0.0509   & 0.6555 & 0.3084 & - 
	\\    
        Text-Monkey~\cite{liu2024textmonkey} & 10B
	& 0.7331  &  0.0371 &  0.4551   & 0.1142  & - 
	\\
        InternVL 1.5~\cite{chen2024far} & 26B
	& 0.7590  & 0.0222  & 0.3601  & 0.4529 &  0.3577
	\\
          InternVL 2 & 26B
	& 0.8855  & 0.0176  & 0.2320  & 0.5019 & 0.4125 	\\
	GPT-4V & N/A & \textbf{0.9821}
	& \underline{0.0096}  &   \textbf{0.0431}  & \underline{0.6101} & \underline{0.6501}
        \\
        GPT-4o & N/A & \underline{0.9761}
	& \textbf{0.0095}   &  \underline{0.0654}  & \textbf{0.7183} & \textbf{0.6762}    \\
    \bottomrule
	\end{tabular}
	}
	\label{tab:compared_sota}
\vspace{-0.30cm}
\end{table*}

\noindent \textbf{Compared Baselines.} We select various models as baselines for different tasks to provide comprehensive comparisons. Specifically, various multi-modal language models, \textit{e.g.}, QWen-VL~\citep{bai2023qwen}, CogAgent~\citep{hong2023cogagent}, DocOwl-1.5~\citep{hu2024mplug}, Text-Monkey~\citep{liu2024textmonkey}, IntenVL~1.5~\citep{chen2024far}, and GPT-4V~\citep{2023gpt4V} are tested on document classification, visual grounding, open-ended single-page QA and multi-page QA tasks. For the Document Layout Detection task, we compare DocXChain~\citep{yao2023docxchain} and YOLOv8~\citep{yolov8}. Additionally, we employ Mathpix, a representative commercial software for mathematical formula transformation, as the compared method for the Document Transformation task, including Equation-to-\LaTeX\ and Table-to-\LaTeX\ transformations.

\noindent \textbf{Implementation Details.} 
\label{sec:implementation}
We utilize a combination of document images and instruction prompts as the input. Note that all tasks use a single-page document image as the input, except for the multi-page QA task, which contains at least two consecutive pages of the document. Besides, the multi-page QA task can only be evaluated on the models that support multi-image inputs. For the layout detection task, which uses the single-page document image as input, we use YOLOv8~\citep{yolov8} as the training baseline, trained for 30 epochs with the AdamW optimizer~\citep{Adamw}, with a learning rate of 0.01. For Equation-to-\LaTeX\ and Table-to-\LaTeX\ tasks, we first use the layout annotations to crop out different modalities, \textit{e.g.}, Table, Equation, \textit{etc}., from the original images. We then employ the same model structure as Pix2Struct-B (0.2B parameters)~\citep{lee2023pix2struct} to perform the fine-tuning on DocGenome-train, resulting in EqVLM-B and TableVLM-B. The fine-tuning process lasts for 30 epochs on 64 NVIDIA A100 80G GPUs, with an initial learning rate of $0.00005$ and a weight decay of $0.01$.

\subsection{Performance on DocGenome-test}

We evaluate the performance of several state-of-the-art multi-modal large language models on the proposed DocGenome-test, covering document classification, visual grounding, and both single-page and multi-page QA tasks. As shown in Table \ref{tab:compared_sota}, among the tested models, GPT-4V~\citep{2023gpt4V} achieves the highest classification accuracy with 98.0\% Top-1 Acc, while QWen-VL~\citep{bai2023qwen} and InternVL 1.5~\citep{chen2024far} also show competitive results with 82.4\% and 75.9\% accuracy, respectively. For the visual grounding task, GPT4V showcases the best performance in the Title OCR Grounding task with the lowest Edit Distance of $0.0104$, while InternVL 1.5 outperforms other models in the Abstract OCR Grounding task with the lowest Edit Distance of $0.3601$. In the single-page QA task, GPT-4V attains the highest GPT-acc score of 61.0\%, indicating its superior ability to handle document-based QA tasks. For the multi-page QA task, GPT-4V again leads with a GPT-acc score of 65.0\%, further demonstrating its robustness in handling multi-page document queries.

\begin{table*}[!t]
\vspace{-0.25cm}
    \small
	\centering	
    \setlength\tabcolsep{4pt}
    \caption
	{
	\small	
        Experiments on scaling up the data using the DocGenome-train, with the resulting models evaluated on document layout detection task. We fine-tune YOLOv8~\citep{yolov8} model using the DocGenome-train with different amounts of training data.
	}
    \vspace{-0.15cm}
	\resizebox{1\textwidth}{!}{
	\begin{tabular}	{c |  c  | c  | c  c  c c c c c }
	\toprule	 	
        Model & Training Data Amount & mAP@0.5:0.95$\uparrow$&  Title& Text& Figure & Caption& Equation &  Table&  Footnote\\
	\midrule  
    \multicolumn{10}{l}{\textit{~~\textbf{Layout detection task on DocGenome-test} }}\\ 
        DocXChain~\cite{yao2023docxchain} & N/A &53.20
	& 49.21 & 79.22 & 43.85  & 48.18 & 49.36 & 72.79 & 29.79
	\\
    \midrule  
         YOLOv8~\cite{yolov8} & 7K & 77.47
	&  71.79 & 92.48 & 76.29 & 86.56 & 80.65 & 85.81  & 48.43
    \\
            YOLOv8~\cite{yolov8} & 70K & 89.42
	&  83.46 & 95.56 & 86.36 & 94.92 & 90.13 & 92.77  & 82.72
    \\
        YOLOv8~\cite{yolov8} & 700K & \textbf{91.37}
	&  \textbf{86.05} &  \textbf{95.96} & \textbf{88.46} & \textbf{95.71} & \textbf{93.06} & \textbf{93.77}  &  \textbf{86.52}
    \\
	\bottomrule
	\end{tabular}
	}
	\label{tab:layout_scale_up}
\end{table*}	
\begin{table*}[!t]
\vspace{-0.10cm}
    \small
    \caption
	{
	\small	
        Experiments on scaling up the data using the DocGenome-train, with the resulting models evaluated on equation and table transformation tasks. EqVLM-B and TableVLM-B mean that we train a visual encoder and a text decoder using the DocGenome-train for the equation and table transformation task, respectively.
	}
    \vspace{-0.20cm}
	\centering	
    \setlength\tabcolsep{4pt}
	\resizebox{0.98\textwidth}{!}{
	\begin{tabular}	{l l |  c  c  c  c }
	\toprule	 	
	 Model & Training Data Amount &  Edit Distance$\downarrow$ & Jaccard Similarity$\uparrow$ & Cosine Similarity$\uparrow$  & BLEU$\uparrow$  \\
	\midrule  
    \multicolumn{6}{l}{\textit{~~\textbf{Equation-to-LaTeX task on DocGenome-test} }} \\ 
	Mathpix\tablefootnote[3]{The version of the online API we used for evaluation: \textcolor{teal}{\url{https://mathpix.com/equation-to-latex}}. \label{mathpix_eq}} & N/A 
	& 0.4738  & 0.7226  & 0.6045  & 0.4472  
	\\
    EqVLM-B & 10K
    & 0.3781 & 0.8157  & 0.7840  & 0.5165 
    \\
    EqVLM-B & 100K
    & 0.2795  & 0.8505  & 0.8317  & 0.5862
    \\
    \textbf{EqVLM-B} & 1M
    &  \textbf{0.2111} &  \textbf{0.8736} & \textbf{0.8621}  &  \textbf{0.6352}
    \\
    \midrule
    \multicolumn{6}{l}{\textit{~~\textbf{Table-to-LaTeX task on DocGenome-test} }} \\
    Mathpix\tablefootnote[4]{Online API we used for evaluation: \textcolor{teal}{\url{https://mathpix.com/table-to-latex}}.}  & N/A 
	& 0.4436  & 0.7730  & 0.5826  & 0.3528 
	\\
    TableVLM-B & 5K
    & 0.4821  & 0.8158  & 0.7804  & 0.4596 
    \\
    TableVLM-B & 10K
    & 0.4738  & 0.8635  & 0.8187  & 0.4973  
    \\
    TableVLM-B & 100K
    & 0.3091  & 0.8903  & 0.8571  & 0.5340   
    \\
    \textbf{TableVLM-B} & 500K
    & \textbf{0.2223}  & \textbf{0.8997}  & \textbf{0.8800}  & \textbf{0.5552}  
    \\

    \bottomrule
	\end{tabular}
	}
	\label{tab:scaleup}
\vspace{-0.10cm}
\end{table*}	


\subsection{Effectiveness of DocGenome-train}

To validate the effectiveness of the proposed DocGenome-train, we further conduct experiments on scaling up the training data using the DocGenome-train dataset, evaluating the performance improvements of different tasks, \textit{e.g.,} layout detection and document transformation tasks.

Specifically, for the layout detection task, we present the evaluation performance of YOLOv8~\citep{yolov8} under three different training scales in Table~\ref{tab:layout_scale_up}. It shows that the model's layout detection capacity continually and significantly improves by increasing the training data volume. Regarding the per-attribute performance improvement, the most significant benefit is observed for ``Footnote'' attribute, which increases from 48.43\% to 86.52\% mAP after scaling up the training data from 7K to 700K. Compared with DocXChain~\citep{yao2023docxchain} that only supports the annotation of seven attributes, our trained YOLOv8 consistently outperforms it in seven attributes, validating the effectiveness of the DocGenome-train.

As illustrated in Table~\ref{tab:scaleup}, for the document transformation task, we conduct similar experiments on Equation-to-\LaTeX\ task and Table-to-\LaTeX\ task, respectively. In these two tasks, we further explore different scaling up settings, with the observation that both tasks benefits the most from scaling up training data from 10K to 100K. Additionally, considering that Edit Distance is more reliable and rigorous to evaluate the similarity, we can observe that the Table-to-\LaTeX\ task has the potential to improve more than the Equation-to-\LaTeX\ task by continuous scaling up. This is because the performance improvement between 100K and 500K training data for TableVLM-B largely exceeds the improvement between 100K and 1M training data for EqVLM-B as shown in Table~\ref{tab:scaleup}.

\begin{table*}[!t]
\vspace{-0.25cm}
    \small
	\centering	
    \caption
	{
	\small	
        Comparisons with state-of-the-art tools on Out-Of-Distribution (OOD) data, where Mathpix is a closed-source commercial software that requires a subscription, while ours is an open-source and free tool.
	}
    \vspace{-0.25cm}
	\resizebox{0.93\textwidth}{!}{
	\begin{tabular}	{c  c | c  c  c c c c c }
	\toprule	 	
        Model  & mAP@0.5:0.95$\uparrow$  &  Title & Text & Figure   & Caption  & Equation  &  Table &  Footnote\\
	\midrule  
    \multicolumn{9}{l}{\textit{~~\textbf{Layout detection task on Human-annotated data}}}\\ 
        DocXChain~\cite{yao2023docxchain} & 37.99
	& 32.53 & 59.00 & \textbf{67.17} & 38.71  & 12.98 & 38.99 & 16.54
	\\
        YOLOv8~\cite{yolov8} & \textbf{50.15} 
	& \textbf{42.59}  &  \textbf{64.87}   & 56.65 & \textbf{64.51} & \textbf{47.14} &  \textbf{47.08}  &   \textbf{28.21}
    \\
    \midrule
	\end{tabular}
	}
	\label{tab:ood}
\end{table*}	

\begin{table*}[!t]
\vspace{-0.25cm}
    \small
	\centering	
    \setlength\tabcolsep{4pt}
	\resizebox{0.93\textwidth}{!}{
	\begin{tabular}	{l l |  c  c  c  c }
	Model &  &Edit Distance$\downarrow$ & Jaccard Similarity$\uparrow$ & Cosine Similarity$\uparrow$  & BLEU$\uparrow$  \\
	\midrule  
    \multicolumn{2}{l}{\textit{~~\textbf{Equation-to-LaTex task on Sci-Hub data} }}\\ 
    Mathpix\textsuperscript{\ref{mathpix_eq}}. & &  \textbf{0.4873}  &  \textbf{0.7437}  &  \textbf{0.7295}  &  \textbf{0.1137}  
	\\
    EqVLM-B
    & & 0.6627  & 0.6303  & 0.5726  & 0.0602 
    \\
    \bottomrule
	\end{tabular}
	}
	\label{tab:equation}
\vspace{-0.35cm}
\end{table*}

\subsection{Further Discussions}

\noindent\textbf{Generalization on Out-Of-Distribution (OOD) Data.} We discuss the generalization ability of models trained on our DocGenome-train to OOD data. Specifically, we conduct experiments on human-annotated data for the layout detection task and Scihub data for the Equation-to-\LaTeX\ task. As shown in Table \ref{tab:ood}, for the layout detection task, YOLOv8~\citep{yolov8} trained using DocGenome-train presents better generalization ability than DocXChain on human-annotated data. Regarding the Equation-to-\LaTeX\ task, although the performance of EqVLM-B declines on OOD data (Scihub data), it still maintains relatively strong results with an Edit Distance of 0.6627. Considering that Mathpix is a closed-source tool with potential exposure to various data distributions in its commercial usage, it is natural that our trained model performs relatively worse than Mathpix in the OOD data.

\noindent \textbf{Potential Applications of DocDenome.} 
1) Conducting document transformation task for more modality types: DocGenome includes various types of data within scientific documents, such as Charts, Equations, Tables, Algorithms, Lists, Codes, and Footnotes, \textit{etc}. For this paper, we study the document transformation using only two types of modalities: Table-to-\LaTeX\ and Equation-to-\LaTeX. Similarly, we can also train a model (image-encoder followed by a text-decoder) that can address the Algorithm-to-\LaTeX\ or List-to-\LaTeX\ transformation task, \textit{etc} using DocGenome.

2) Performing document-level tasks with entity relations: DocGenome contains the logical relationships between component units,  we can input different component units to examine the model's understanding of long-range contextual relationships.

3) Conducting document OCR task on any page at any location: the layout annotations of DocGenome are very comprehensive, covering almost all locations in the document, and DocGenome has the ground truth text of the entire document. Therefore, we can use the layout information and text information to perform OCR tasks on any page at any location, not just the title and abstract regions, which further examines both the OCR capability and the visual grounding capability of the model.
\vspace{-0.25cm}


\section{Conclusion}


In this paper, we introduced DocGenome, a large-scale, structured, multi-task, and multi-modal dataset for scientific documents. We constructed DocGenome using DocParser, our developed auto-labeling pipeline, to extract structured attributes and relationships between units. DocGenome's comprehensive task coverage, logicality, diversity, and correctness make it a valuable resource for training models related to scientific documents and evaluating the capabilities of such large models.

\section*{Acknowledgement}

The research was supported by the National Key R\&D Program of China (Grant No. 2022ZD0160104), the Science and Technology Commission of Shanghai Municipality (Grant No. 22DZ1100102), and Shanghai Rising Star Program (Grant No. 23QD1401000).





{\small
\normalem
\bibliographystyle{plainnat}
\bibliography{main}
}

\clearpage
\newpage
\setcounter{page}{1}
\setcounter{equation}{0}
\setcounter{figure}{0}
\setcounter{table}{0}
\appendix
\renewcommand\thefigure{A.\arabic{figure}}
\renewcommand\theequation{A.\arabic{equation}}
\renewcommand\thetable{A.\arabic{table}}

\section{Overview of Appendix}

We provide more information on our benchmark and further experiment details from the following aspects:

\begin{itemize}
    \item Sec.~\ref{app:limitation}: Limitations and Dataset Accessibility.
    \begin{itemize}
        \item Sec.~\ref{sec:limit}: Limitations.
        \item Sec.~\ref{sec:data_access}: Dataset Accessibility.
    \end{itemize}
    \item Sec.~\ref{app:anno_explain}: Annotation Explanations.
    \item Sec. \ref{app:distribution}: More Statistical Distributions of DocGenome.
    \item Sec.~\ref{app:check_details}: Details of Quality Assurance.
    \item Sec.~\ref{app:gpt_eval}: Prompt Design for GPT-acc.
    \item Sec.~\ref{app:visualization}: Annotation Examples in DocGenome.
    \item Sec.~\ref{app:visual_task}: Task Examples in DocGenome-test.
\end{itemize}

\section{Limitations and Dataset Accessibility}
\label{app:limitation}
\vspace{-0.15cm}

\subsection{Limitations}
\label{sec:limit}
\vspace{-0.15cm}

The purpose of our DocGenome is to build a comprehensive scientific document dataset, promoting the development of intelligent document processing and effective evaluation of MLLMs in document understanding tasks. Although our DocGenome provides annotations for 6 categories of entity relationships, exploring the impact of these entity relationship annotations on large models' understanding of scientific documents is highly meaningful. For future works, we will explore the role of the entity relationships in understanding scientific documents. 

\subsection{Dataset Accessibility}
\label{sec:data_access}
\vspace{-0.15cm}

\noindent \textbf{Dataset Documentation:} We have documented our dataset and its intended uses, as required. The website of our dataset is available at the following link: \textcolor{teal}{\url{https://github.com/UniModal4Reasoning/DocGenome}}, which includes metadata, format details, and visualizations. Besides, the download link for the dataset is: \textcolor{teal}{\url{https://drive.google.com/drive/folders/1OIhnuQdIjuSSDc_QL2nP4NwugVDgtItD?usp=sharing}}.

\noindent \textbf{Dataset Statistics and Analyses:} We have conducted extensive data statistics and analyses, along with thorough quality checks including DocGenome-train and DocGenome-test datasets, which are presented in Sec.~\ref{sec:doc_analy} and Sec.~\ref{sec:qa_pair_quality_check}.

\noindent \textbf{Long-term Preservation:} To ensure the long-term preservation of the DocGenome dataset, we have uploaded it to Google Drive\footnote[4]{The download link for the dataset is available at: \textcolor{teal}{\url{https://drive.google.com/drive/folders/1OIhnuQdIjuSSDc_QL2nP4NwugVDgtItD?usp=sharing}}.}. This ensures continuous accessibility to the dataset for an extended duration. Furthermore, we will routinely back up the data and monitor its availability to maintain continued accessibility.

\noindent \textbf{Terms of Use and License:} We have chosen the CC BY 4.0 license for our dataset, as required. This information is included in our paper submission and will also be clearly stated on our dataset website.

\noindent \textbf{A Persistent Dereferenceable Identifier:} We have obtained a DOI for our dataset, referred to as \textcolor{teal}{10.5281/zenodo.11488587}. This persistent dereferenceable identifier ensures long-term accessibility and citability of the dataset.

\noindent \textbf{Discussion of Personally Identifiable Information.} All the scientific documents in our DocGenome are sourced from the arXiv open-access community, where papers are released under the CC license. Besides, the arXiv community ensures that papers uploaded by authors adhere to legal and ethical guidelines, including the protection of personal information and the avoidance of offensive material. Thus, we can confirm that our DocGenome does not contain personally identifiable information or offensive content.

\begin{figure*}[t]
    \centering
    \resizebox{1\linewidth}{!}{\includegraphics{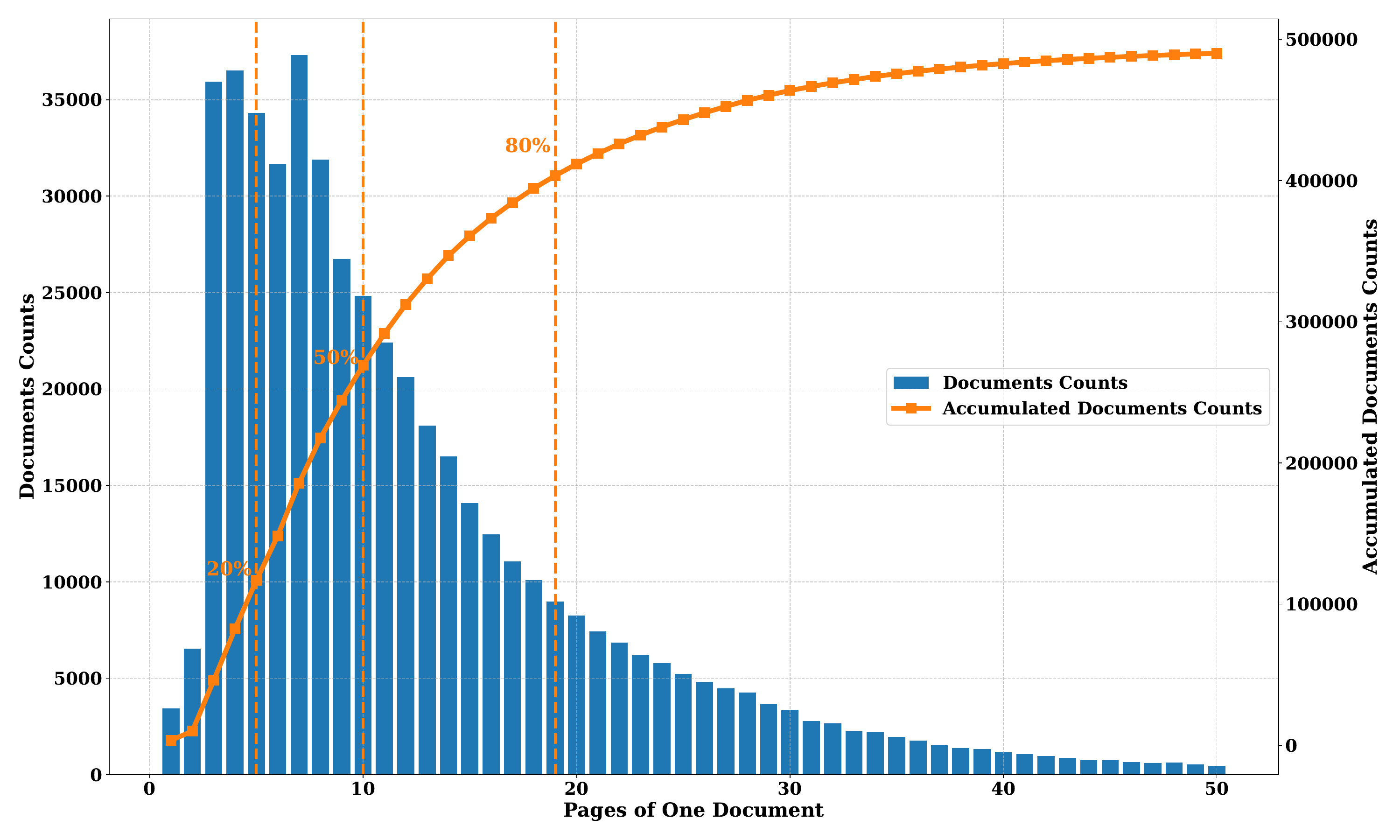}}
    \caption{Page distribution of DocGenome. 20\% of documents are five pages or fewer, 50\% are ten pages or fewer, and 80\% are nineteen pages or fewer.}
    \label{fig:page_distribution}
\vspace{-2pt}
\end{figure*}

\section{Annotation Explanations}
\label{app:anno_explain}
\vspace{-0.15cm}
We provide the annotation details of DocGenome in Table \ref{tab:layout_annotation}, where the index number in the annotation corresponds to the category index in the attribute list.

\begin{table}[t]
\small
\centering
\tablestyle{0.5pt}{1.0}
\setlength\tabcolsep{1pt}
\caption{Category descriptions of the layout annotation performed by our DocParser. Note that we do not use the ``others'' category and the ``reference'' category, and their indices are 6 and 11, respectively.}
\begin{tabular}{y{50}y{65}y{160}}
\toprule
\textbf{Index} & \textbf{Category}  & \textbf{Notes}            \\ \midrule
0  &  Algorithm   &                              \\ 
1  &  Caption     & Titles of Images, Tables, and Algorithms\\ 
2  &  Equation    &                              \\ 
3  &  Figure      &                              \\ 
4  &  Footnote    &                              \\ 
5  &  List        &                              \\ 
7  &  Table       &                              \\ 
8  &  Text        &                              \\ 
9  &  Text-EQ     & Text block with inline equations \\ 
10  &  Title      & Section titles                \\ 
12  &  PaperTitle &                              \\ 
13  &  Code       &                              \\ 
14  &  Abstract   &                              \\ \bottomrule
\end{tabular}
\label{tab:layout_annotation}
\end{table}

\begin{figure*}[t]
    \centering
    \resizebox{1\linewidth}{!}{\includegraphics{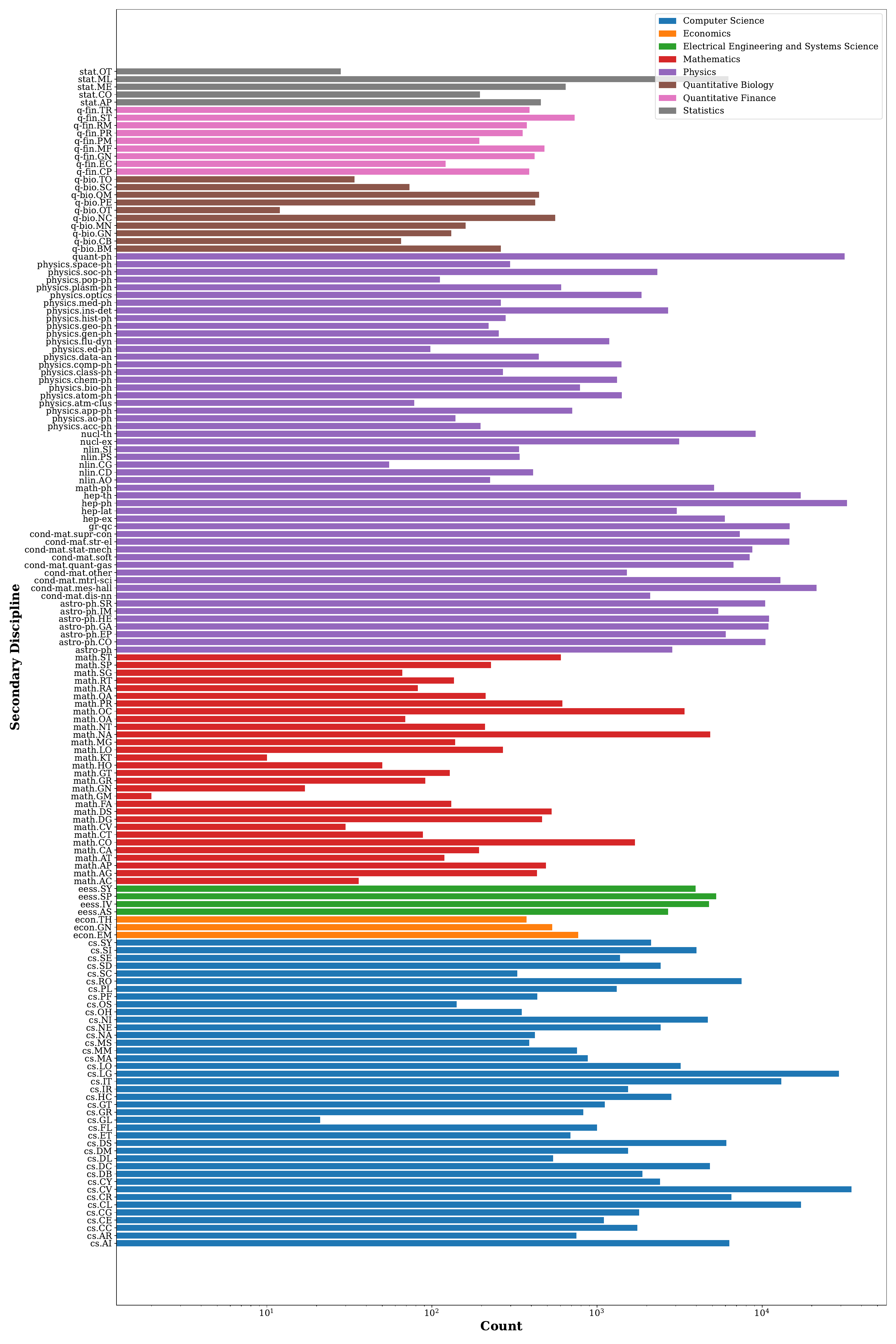}}
    \caption{Distribution of secondary disciplines in our DocGenome. The count on the x-axis represents the number of documents, and documents from the same primary discipline are marked with the same color.} 
    \label{fig:secondary_distribution}
\vspace{-2pt}
\end{figure*}

\section{More Statistical Distributions of DocGenome}
\label{app:distribution}
In addition to the statistical distribution described in Sec. \ref{fig:docgenome_statistic}, we provide more statistical distributions in this section. As shown in Fig. \ref{fig:secondary_distribution}, the sample counts of all secondary disciplines are summarized and marked with different colors, from which it can be observed that the inter-discipline and intra-discipline distributions are both diverse, with Physics, Computer Science, and Mathematics papers occupying the major components of DocGenome.

We also present the page distribution of DocGenome in Fig. \ref{fig:page_distribution}, which indicates the diversity of paper length in DocGenome. Specifically, 50\% papers in DocGenome have nearly or fewer than 10 pages, with 80\% papers having fewer than 19 pages.

\section{Details of Quality Assurance for QA Data}
\label{app:check_details}
\begin{figure*}[t]
    \centering
    \resizebox{0.85\linewidth}{!}{\includegraphics{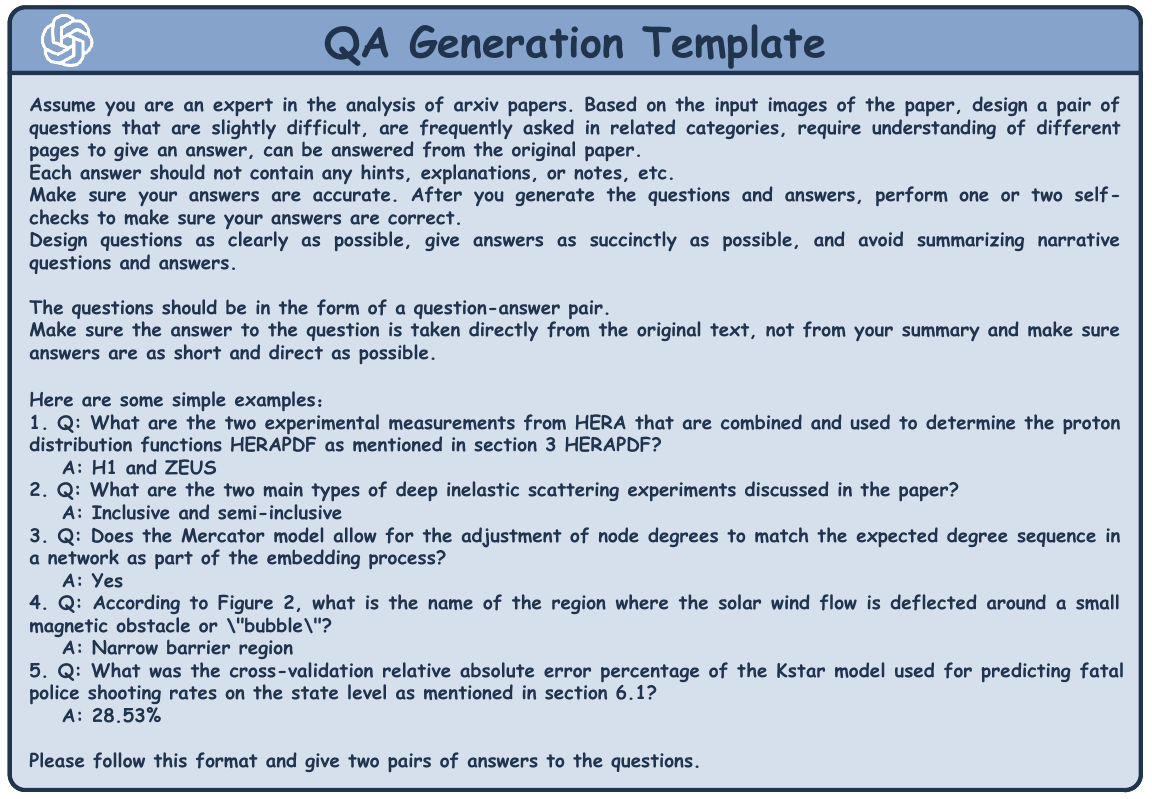}}
    \caption{Template prompts using GPT-4V~\citep{2023gpt4V} for document QA pair generation.}
    \label{fig:qa_gen}
\vspace{-2pt}
\end{figure*}

\noindent\textbf{The QA Generation Details.}
We provide a general prompt template for QA pair generation in Fig.~\ref{fig:qa_gen}. The discipline-specific guidance is imposed to generate the corresponding ground-truth labels to achieve diversity and relevance.

\noindent\textbf{The Quality Checking Details.}
During independent verification by professional faculty members, each judgment was assigned with a confidence value ranging from 0 to 3. The confidence criterion is designed as follows:

\textbf{Confidence 3}: The reviewer is confident that the QA pair is accurate and relevant to the provided paper.

\textbf{Confidence 2}: The reviewer thinks the QA pair is mostly accurate and relevant to the provided paper but is unsure whether it is absolutely correct.

\textbf{Confidence 1}: The reviewer has no idea about the correctness or relevance of the QA pair to the provided paper.

\textbf{Confidence 0}: The reviewer is confident that the QA pair is wrong or irrelevant to the provided paper.

During the cross-verification, the confidence values of the two professional faculty reviewers were compared with the automatically-annotated correctness. The QA pairs with inconsistent results were re-analyzed by the two reviewers and updated to a precise version with consistent confidence.

\section{Prompt Design for GPT-acc}
\label{app:gpt_eval}
\begin{figure*}[t]
    \centering
    \resizebox{0.85\linewidth}{!}{\includegraphics{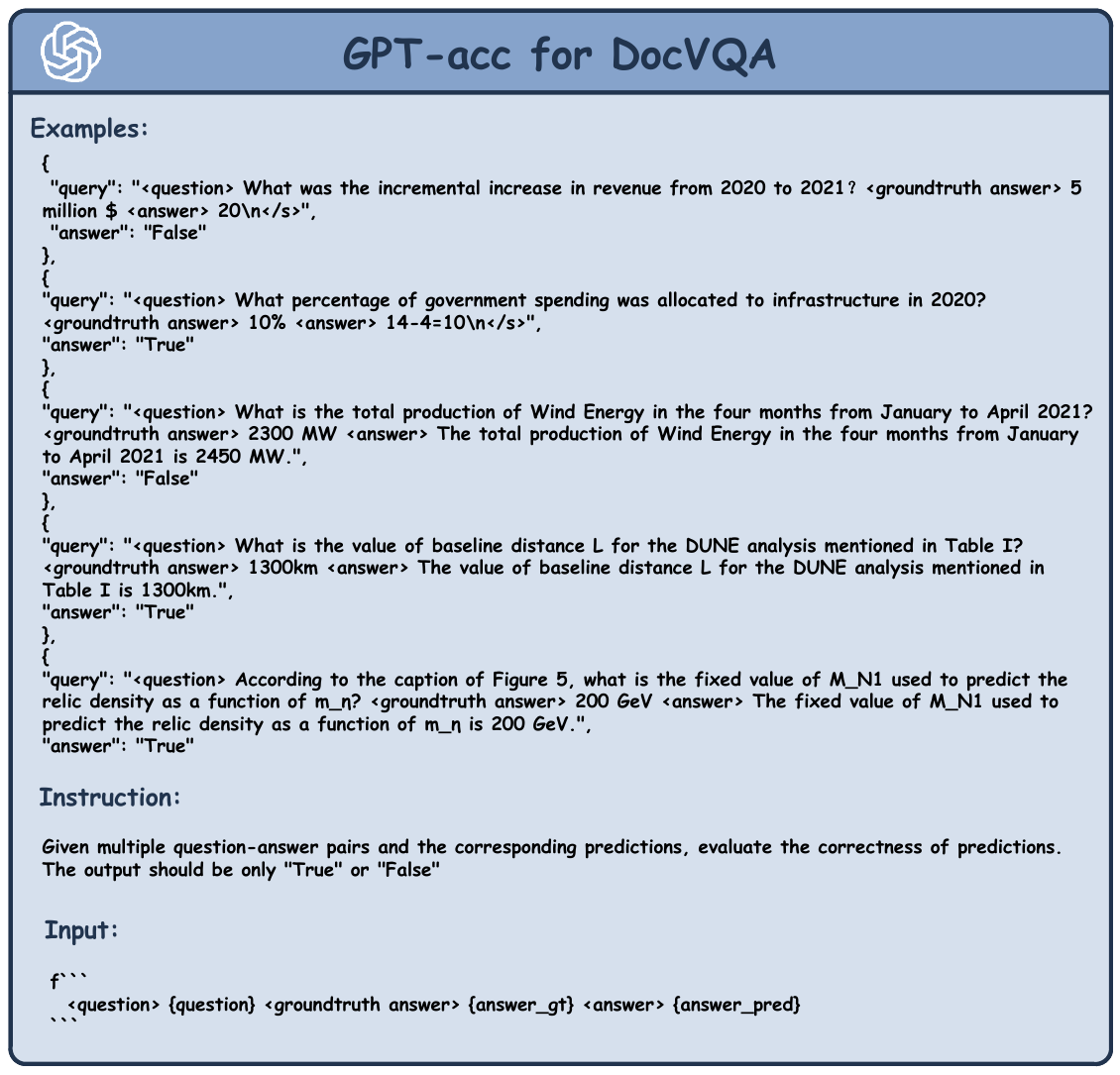}}
    \caption{Detailed prompts in GPT-acc metric for document QA tasks.}
    \label{fig:gpt_acc}
\vspace{-2pt}
\end{figure*}

We adopt GPT-acc as the evaluation metric for the QA tasks. The complete prompts are concluded in Fig.~\ref{fig:gpt_acc}.

\section{Examples in Document-level Annotation from DocGenome}
\label{app:visualization}
\begin{figure*}[t]
    \centering
    \resizebox{1\linewidth}{!}{\includegraphics{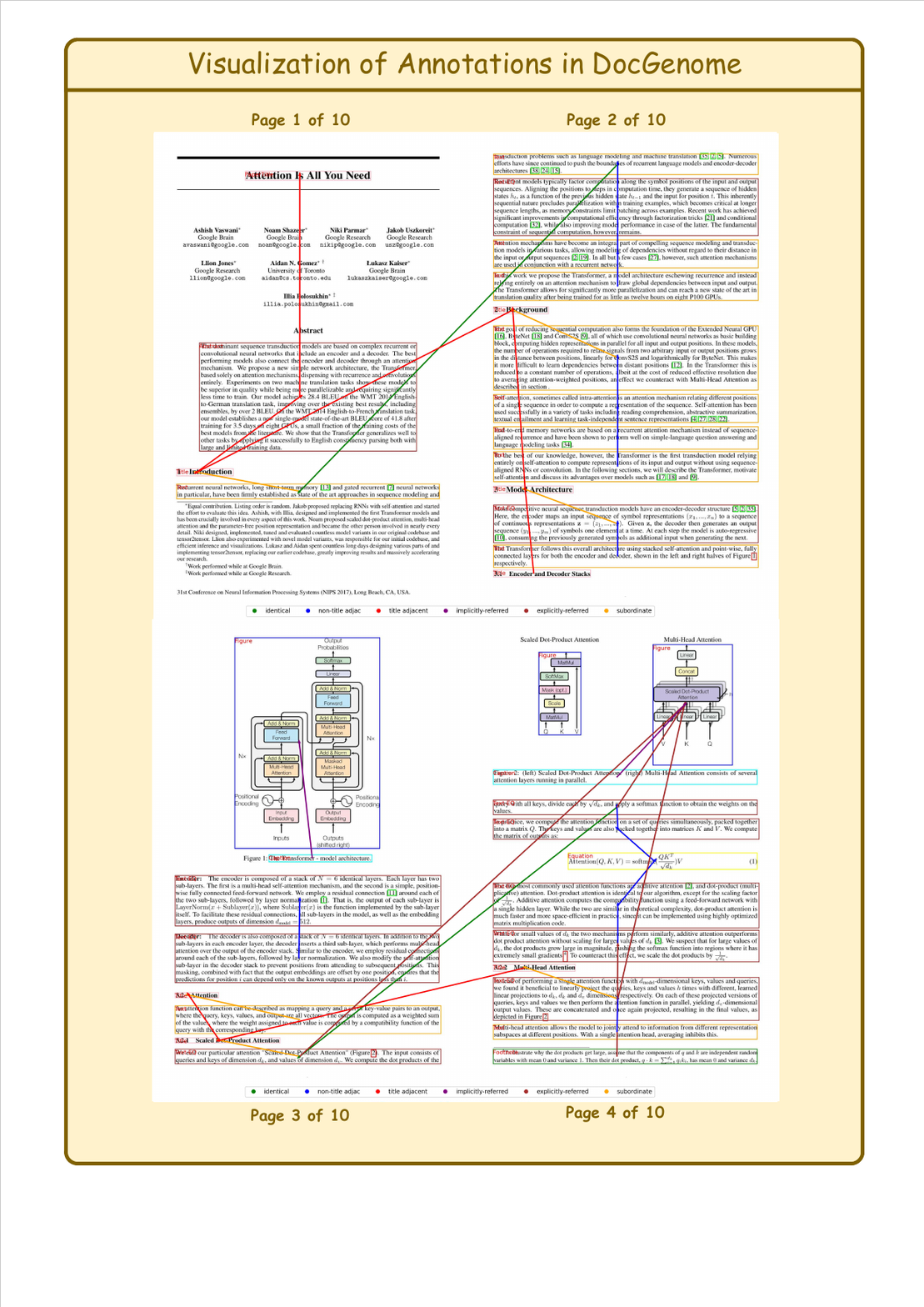}}
    \caption{Annotations of a complete document in DocGenome, taking `\textit{Attention is All Your Need}'~\citep{Vaswani2017AttentionIA} as an example.} 
    \label{fig:example1}
\vspace{-2pt}
\end{figure*}

\begin{figure*}[t]
    \centering
    \resizebox{1\linewidth}{!}{\includegraphics{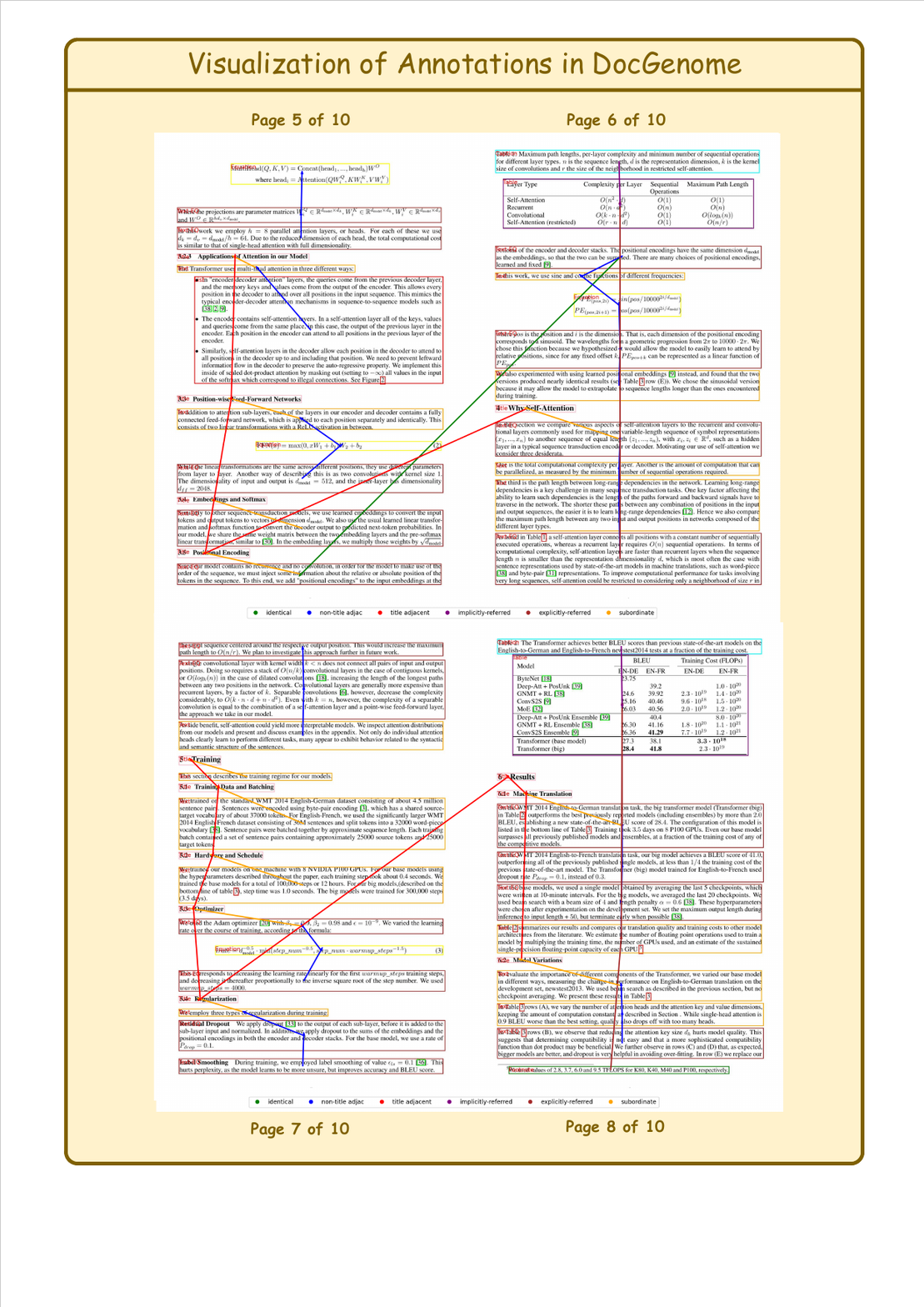}}
    \caption{Annotations of a complete document in DocGenome, taking `\textit{Attention is All Your Need}'~\citep{Vaswani2017AttentionIA} as an example.} 
    \label{fig:example2}
\vspace{-2pt}
\end{figure*}

\begin{figure*}[t]
    \centering
    \resizebox{1\linewidth}{!}{\includegraphics{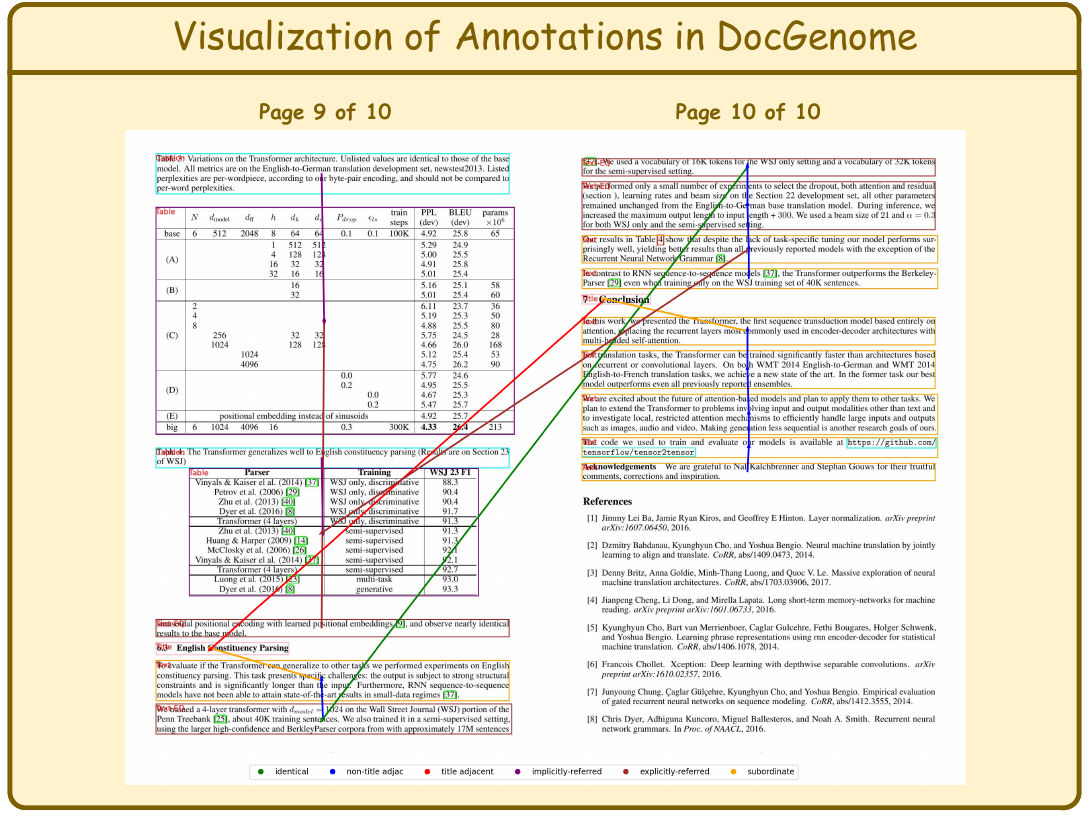}}
    \caption{Annotations of a complete document in DocGenome, taking `\textit{Attention is All Your Need}'~\citep{Vaswani2017AttentionIA} as an example.} 
    \label{fig:example3}
\vspace{-2pt}
\end{figure*}

We present one example in DocGenome in Figs. \ref{fig:example1}, \ref{fig:example2}, and \ref{fig:example3} to visualize the annotations of each page in a whole document~\citep{Vaswani2017AttentionIA}. The blocks marked with different colors refer to different attributes of component units and the arrows with different colors denote different relations between units.

\section{Examples of Tasks in DocGenome-test}
\label{app:visual_task}
\begin{figure*}[t]
    \centering
    \resizebox{1\linewidth}{!}{\includegraphics{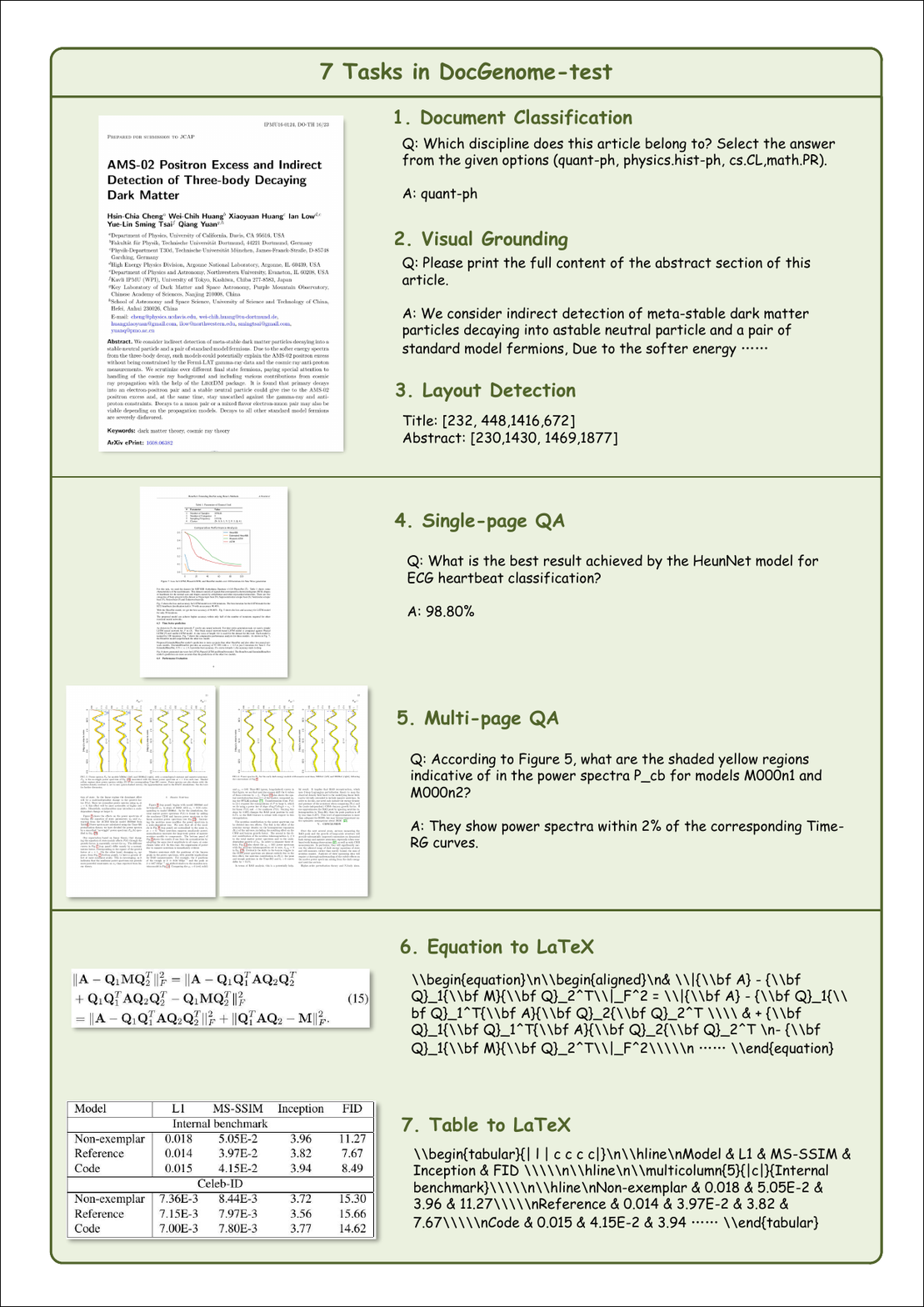}}
    \caption{Visualization examples of 7 tasks in DocGenome-test.} 
    \label{fig:visual_task}
\vspace{-2pt}
\end{figure*}

We provide visual demonstrations in Fig.~\ref{fig:visual_task} for all 7 tasks in DocGenome-test, including document classification, visual grounding, open-ended single-page and multi-page QA tasks, document layout detection, Equation-to-\LaTeX\ transformation, and Table-to-\LaTeX\ transformation. 

\end{document}